\documentclass{article}

\usepackage[table]{xcolor}
\usepackage[final]{corl_2026} % Camera-ready version.

\makeatletter
\renewcommand{\@conferencelocation}{Austin, Texas, USA}
\makeatother

\usepackage{amsmath}
\usepackage{amssymb}
\usepackage{amsfonts}
\usepackage{graphicx}
\usepackage{caption}
\usepackage{booktabs}
\usepackage{multirow}
\usepackage{float}
\usepackage{tabularx}
\usepackage{ragged2e}
\usepackage{wrapfig}

\title{TTT-Parkour: Rapid Test-Time Training for Perceptive Legged Humanoid Parkour}

\author{%
Shaoting Zhu$^{1,2,*}$ \quad Baijun Ye$^{1,2,*}$ \quad Jiaxuan Wang$^{1,\dagger}$ \quad Jiakang Chen$^{1,\dagger}$\\
\textbf{Ziwen Zhuang$^{1,2}$ \quad Linzhan Mou$^{3}$ \quad Runhan Huang$^{1}$ \quad Hang Zhao$^{1,2,\ddagger}$}\\
$^{1}$Tsinghua University \quad $^{2}$Shanghai Qi Zhi Institute \quad $^{3}$Princeton University\\
$^{*}$Equal contribution \quad $^{\dagger}$Equal contribution \quad $^{\ddagger}$Corresponding author
}

\hypersetup{
  pdftitle={TTT-Parkour: Rapid Test-Time Training for Perceptive Legged Humanoid Parkour},
  pdfauthor={Shaoting Zhu, Baijun Ye, Jiaxuan Wang, Jiakang Chen, Ziwen Zhuang, Linzhan Mou, Runhan Huang, Hang Zhao}
}

\begin{document}
\maketitle

\begin{abstract}
Achieving highly dynamic parkour with humanoid robots on complex terrains unseen during the pre-training stage remains a challenge in robotics. Although general locomotion policies demonstrate capabilities across broad terrain distributions, they often struggle with arbitrary and highly challenging environments. To overcome this limitation, we propose a real-to-sim-to-real framework for rapid test-time training (TTT): environment-specific policy fine-tuning in simulation before deployment on a newly encountered terrain. We adopt a two-stage end-to-end learning paradigm: a policy is first pre-trained on diverse procedurally generated terrains, followed by rapid fine-tuning on high-fidelity meshes reconstructed from real-world captures. Specifically, we develop an efficient, feed-forward, and high-fidelity geometry reconstruction pipeline using RGB-D inputs, ensuring both speed and quality during test-time training. We demonstrate that \textit{TTT-Parkour} empowers humanoid robots to master complex obstacles, including wedges, stakes, boxes, trapezoids, and narrow beams. The entire pipeline, including capture, reconstruction, and test-time training, takes approximately 10 minutes for most tested terrains. Experiments show improved traversal on the tested terrains after fine-tuning, with direct sim-to-real deployment and no further policy updates on the robot. Project page: \url{https://ttt-parkour.github.io/}.

\end{abstract}


\section{Introduction}

% Keep teaser text wrapping on the first page.
\begin{wrapfigure}[17]{r}{0.5\textwidth}
  \vspace{-4mm}
  \centering
  \includegraphics[width=\linewidth]{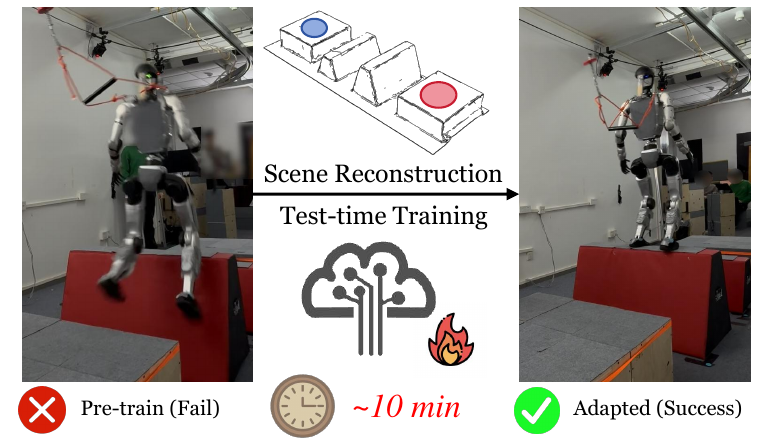}
  \caption{Rapid test-time training on terrains unseen during pre-training. By reconstructing the scene and fine-tuning in simulation, our framework enables the robot to master challenging obstacles in approximately 10 minutes for most tested terrains, turning failure (left) into success (right).}
  \label{fig:teaser}
\end{wrapfigure}

Recent advancements in deep reinforcement learning have fundamentally revolutionized humanoid locomotion control, enabling robots to demonstrate robust mobility in diverse real-world environments \cite{he2025attention, rudin2025parkour, zhu2026hiking}. By leveraging massive parallel simulation and sim-to-real transfer techniques \cite{mittal2025isaac, makoviychuk2021isaac}, humanoid robots can now traverse unstructured terrains. 
Although some general policies demonstrate capabilities across broad terrain distributions, they struggle to traverse complex obstacles unseen during pre-training. 
Bridging this gap to achieve true athletic intelligence is critical for the deployment of humanoids in challenging environments.

Procedural terrain generation supports broad pre-training but may leave difficult real-world configurations uncovered. Our framework targets these remaining failure cases through rapid adaptation to a reconstructed target environment.

Moreover, manually reproducing realistic geometric features in simulation is labor-intensive. While per-scene optimization methods like NeRF \cite{kerr2022evo, zhou2023nerf, ye2024nerf, byravan2022nerf2real} and 3DGS \cite{lou2025robo, qureshi2025splatsim, ye2025gs, li2024robogsim} excel in view synthesis and extend to physical interaction \cite{escontrela2025gaussgym, zhu2025vr, han2025re}, the overall workflow is often computationally intensive and time-consuming, making it incompatible with the rapid adaptation requirements of test-time training.
Conversely, feed-forward \cite{wang2025vggt, wang2025pi, wu2026rays, keetha2025mapanything} and generative approaches \cite{chen2025sam3d} are faster but often yield scale-ambiguous or distorted geometries on some terrains, leaving them unsuitable for physics simulation in our parkour settings. Thus, generating collision-accurate, simulation-ready meshes within the tight time windows required for rapid adaptation remains a critical bottleneck.

To address these challenges, we introduce \textit{TTT-Parkour}, a real-to-sim-to-real framework designed for rapid humanoid adaptation on challenging terrains, including wedges, stakes, boxes, trapezoids, and narrow beams. Our approach is built upon a two-stage end-to-end perceptive locomotion learning paradigm. We first pre-train a general policy on a diverse set of procedurally generated terrains. Subsequently, we fine-tune the policy in simulation on meshes reconstructed from manually captured real-world terrains before deployment. 
We develop an efficient and high-fidelity geometry reconstruction pipeline using RGB-D input. 
We employ a feed-forward method with automatic scale recovery and frame alignment to directly reconstruct simulation-ready meshes. 
This pipeline enables test-time training of the policy on accurate geometric constraints, significantly accelerating adaptation and mitigating the sim-to-real gap. 
The capture, reconstruction, and test-time training phases complete in approximately 10 minutes for most tested terrains.
The framework builds on established locomotion learning components and provides terrain-specific policies for difficult environments.
In summary, our main contributions are as follows:
\begin{itemize}
    \item We introduce a \textbf{real-to-sim-to-real adaptation pipeline} that combines general-policy pre-training with rapid environment-specific fine-tuning in simulation.
    \item We develop a \textbf{fast, feed-forward, and high-fidelity geometry reconstruction pipeline} to generate simulation-ready meshes from RGB-D inputs, enabling an efficient real-to-sim-to-real parkour workflow.
    \item We demonstrate improved \textbf{humanoid legged parkour performance} over our pre-trained policy on challenging real terrains, and faster specialization than joint multi-terrain fine-tuning.
    
\end{itemize}

\section{Related Work}

\subsection{Perceptive Locomotion}

Integrating exteroceptive perception is crucial for agile locomotion, enabling robots to transition from blind, reactive recovery \cite{kumar2021rma, huang2025moe, gu2024advancing} to proactive obstacle traversal. Traditional map-based approaches typically leverage LiDAR coupled with precise localization to construct elevation maps \cite{he2025attention, long2025learning, wang2025beamdojo, rudin2025parkour} or voxel grids \cite{ben2025gallant}. However, these methods are susceptible to state estimation drift and motion distortion during highly dynamic parkour, making global maps unreliable on extremely challenging terrains. Alternatively, recent works have explored utilizing depth images for policy input \cite{song2025gait, sun2025dpl, duan2024learning, hao2026cref, zhang2026rpl}, retaining intermediate heightmap representations. Following the paradigm of \cite{zhuang2024humanoid, cheng2024extreme, zhu2026hiking, sun2026now, wu2026perceptive}, we employ a forward depth camera to train a policy end-to-end. This approach is robust during high-speed traversal.

\subsection{Fast Adaptation}
Test-time adaptation \cite{xiao2024beyond, sun2020test, behrouz2024titans, bagatella2025test} mitigates out-of-distribution shifts via lightweight online fine-tuning, with gradient-based extensions in foundation models \cite{behrouz2024titans, akyurek2024surprising}. In robot learning, fast adaptation at test time has been extensively explored in recent years: Diffusion-based controllers \cite{janner2022planning, liao2025beyondmimic, huang2025flexible} incorporate gradients of explicit reward and cost functions as guidance at test time to optimize for desired behavioral outcomes. While this class of methods offers flexibility, the generated trajectories may be pulled away from the in-distribution data manifold by gradient updates, resulting in performance collapse.
Other studies use generative models for test-time adaptation to novel environments and embodiments \cite{chen2024mirage}.
Our framework applies environment-specific fine-tuning in a reconstructed simulator before deployment to complement generalist locomotion on difficult terrains.

\subsection{Scene Reconstruction}

High-fidelity reconstruction is critical for enhancing simulation realism and bridging the sim-to-real gap. While per-scene optimization methods like NeRF \cite{kerr2022evo, zhou2023nerf, ye2024nerf, byravan2022nerf2real} and 3DGS \cite{lou2025robo, qureshi2025splatsim, ye2025gs, li2024robogsim, yang2024gs, zhang2024gs, zheng2024gaussiangrasper, li2025scenesplat, li2025chorus} excel in visual synthesis, and recent extensions extract meshes for physical interaction \cite{escontrela2025gaussgym, zhu2025vr, han2025re, torne2024reconciling}, their reliance on offline, multi-stage processes (e.g., COLMAP or iterative optimization) \cite{zhu2025vr, lou2025robo} limits the speed of environment-specific adaptation.
Conversely, recent feed-forward techniques \cite{wang2025vggt, wang2025pi, keetha2025mapanything} bypass optimization but suffer from scale ambiguity or significant metric discrepancies. Recent work has explored unified physical 3D generation for simulation-ready assets \cite{cao2026physx}, but parkour environments involve long-span terrains with complex layouts and occlusions, so single-image generation \cite{chen2025sam3d} often yields severe geometric distortions unsuitable for physics simulation.
To address this, we introduce an efficient pipeline integrating feed-forward reconstruction with automatic scale and frame alignment, producing \textit{simulation-ready} meshes with the speed and fidelity required for test-time parkour training.

\section{Method}

\subsection{Problem Definition}

We define the task as traversing a series of discrete platforms $\mathcal{P} = \{p_{\text{start}}, p_1, \dots,p_{n}, p_{\text{end}}\}$ elevated above the ground $\mathcal{G}_{\text{ground}}$. The robot aims to travel from $p_{\text{start}}$ to $p_{\text{end}}$ following a fixed forward velocity command without an explicit angular velocity command. This parkour setting is restricted to lower-body locomotion: the robot must complete traversal using legged motion and foot-terrain contacts only, without using its hands or upper body to touch or support itself on the terrain. To ensure valid traversal, contact with the ground plane $\mathcal{G}_{\text{ground}}$ is treated as a failure state, preventing the robot from bypassing obstacles by moving on the ground. The terrains consist of geometric primitives with limited contact areas (e.g., wedges, stakes, boxes, trapezoids, and narrow beams), requiring the policy to maintain stability through precise foothold selection.

\subsection{Policy Pre-training}

\begin{figure}[t]
  \centering
  \includegraphics[width=\textwidth]{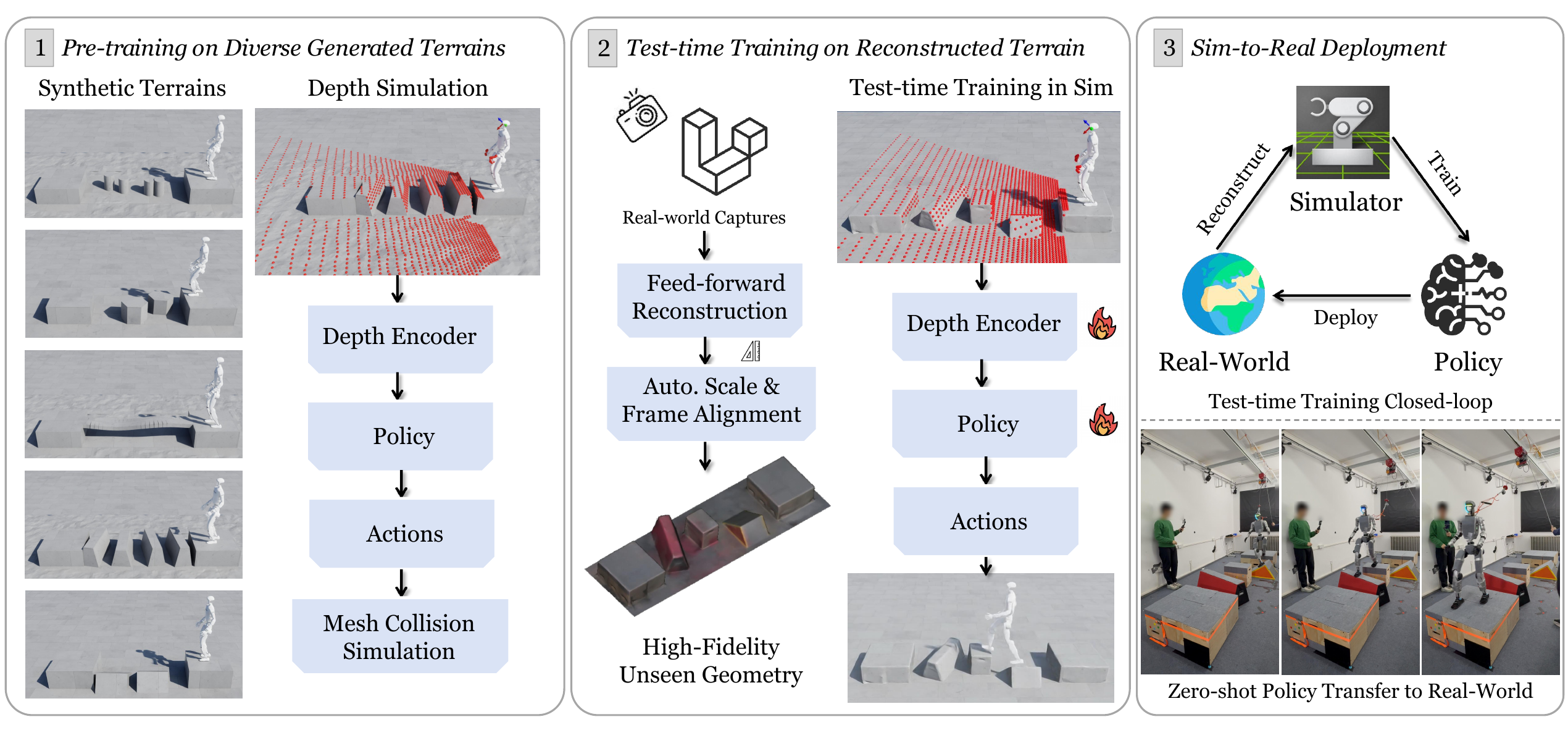}
  \vspace{-4mm}
  \caption{\textbf{TTT-Parkour}. Our framework consists of three stages: (1) Pre-training: A general policy is pre-trained on diverse procedurally generated terrains to learn robust locomotion primitives. (2) Test-time Training (TTT): We reconstruct high-fidelity and simulation-ready meshes from real-world captures using feed-forward reconstruction with automatic scale recovery and frame alignment. The policy is then rapidly fine-tuned on these specific terrains in simulation. (3) Sim-to-Real Deployment: The adapted policy is deployed on the target terrain without further policy updates on the robot.}
  \vspace{-5mm}
  \label{fig:overview}
\end{figure}

We formulate the parkour task as a POMDP and optimize with PPO \cite{schulman2017proximal}. The actor receives a history of $h$ proprioceptive states (angular velocity, projected gravity, velocity commands, joint positions/velocities, last action) concatenated with a strided depth history $\mathcal{H}_t = \{\mathbf{I}_{t-k\ell}\}_{k=0}^{m-1}$, with stochastic noise injected during training to bridge the sim-to-real gap. A CNN-based visual encoder and projection MLP encode the depth history into a visual embedding, which is concatenated with the proprioceptive observations and fed into an actor-critic network. An asymmetric critic additionally observes noise-free states and base linear velocity. The policy outputs target joint positions $\mathbf{a}_t \in \mathbb{R}^{29}$ executed via PD control ($\boldsymbol{\tau}_t = k_p(\mathbf{a}_t - \mathbf{q}_t) - k_d\dot{\mathbf{q}}_t$) with gains from \cite{liao2025beyondmimic}. Episodes terminate on timeout, stalling at the start, ground contact, or excessive base tilt. The reward comprises task tracking, regularization, safety, and AMP-style \cite{peng2021amp} terms; full specifications are in \autoref{app:policy_details}.

\subsection{Efficient Geometry Reconstruction}
\begin{figure}[t]
  \centering
  \includegraphics[width=\linewidth]{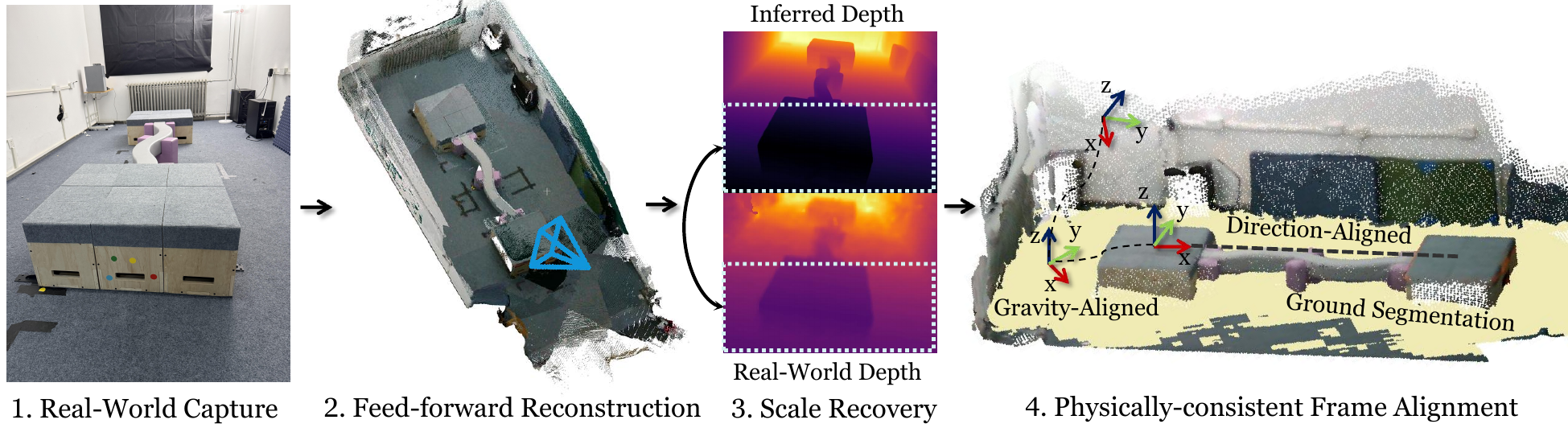}
  \vspace{-4mm}
  \caption{\textbf{Efficient Geometry Reconstruction.} Our pipeline consists of four stages: (1) Real-World Capture. (2) Feed-forward Reconstruction provides initial scene geometry. (3) Scale Recovery corrects metric scale discrepancies by aligning inferred depth with sensor depth. (4) Physically-consistent Frame Alignment registers the terrain to the simulation coordinate system by aligning the $z$-axis with gravity and the $x$-axis with the traversal direction using 3D semantic segmentation.}
  \vspace{-4mm}
  \label{fig:geo_pipe}
\end{figure}

To facilitate rapid test-time training on terrains unseen during pre-training, we introduce an efficient, automated, and high-fidelity reconstruction pipeline that integrates feed-forward reconstruction with automatic scale recovery and frame alignment, shown in \autoref{fig:geo_pipe}.
Formally, we define the reconstruction problem as transforming raw real-world captures $\mathcal{P}_{1:n}$ into a \textit{simulation-ready} mesh $\mathcal{M}$ that is strictly aligned with both the gravity axis $\mathbf{g}$ and the start-to-goal traversal direction $\mathbf{d}$.

\textbf{Feed-forward Terrain Reconstruction:}
We initiate the reconstruction process by employing a feed-forward model \cite{wang2025pi} that takes RGB sequences as input to reconstruct a scale-ambiguous point cloud. Subsequently, we apply Screened Poisson surface reconstruction \cite{kazhdan2013screened} to recover the mesh.

\textbf{Scale Recovery:} 
Existing RGB-only and metric feed-forward approaches \cite{wang2025pi,keetha2025mapanything} are unreliable for predicting precise absolute scale on some terrains, as shown in \autoref{tab:scale_errors}. 
This ambiguity is critical because standard scale randomization (e.g., $s \in [0.9, 1.1]$) is insufficient to compensate for arbitrary scale biases on terrains unseen during pre-training. For instance, if the scale is significantly under-estimated on geometry-constrained terrains like stakes or narrow beams, the effective contact area in the simulator will become infeasibly small. This makes the task physically intractable, preventing policy convergence.

To ensure precise alignment, we calculate a scaling factor by aligning the predicted depth from the feed-forward model to the metric depth from the RGB-D camera. Specifically, we compute the ratio of median depth values derived from the \textit{lower half} of the depth images. This region-of-interest selection effectively focuses on the terrain while mitigating interference from distant background outliers. This alignment step is essential, as it minimizes the sim-to-real gap and ensures that any residual scale deviation falls within the tractable bounds of standard domain randomization.

\textbf{Coordinate System Alignment:}
Given the scaled point cloud and reconstructed mesh, our goal is to register the terrain into a \textit{physically consistent world frame}. In this frame, the origin is anchored at the centroid of the start platform $p_{\text{start}}$, while the $z$-axis and $x$-axis are aligned with the gravity and the intended traversal direction, respectively.

We assume the physical ground plane is orthogonal to the gravity vector. To estimate this plane robustly, we first utilize a 3D segmentation model \cite{zhang2025concerto} to extract points semantically labeled as ground. We apply RANSAC \cite{fischler1981random} to robustly filter outliers from the segmented ground points, followed by PCA to precisely estimate the surface normal $\mathbf{n}$. The entire scene is then rotated to align $\mathbf{n}$ with the $z$-axis in simulation.

To align the traversal direction, we identify the centroids of the starting ($p_{\text{start}}$) and ending ($p_{\text{end}}$) platforms derived from the semantic segmentation. The scene is then rotated around the $z$-axis such that the vector connecting these centroids aligns with the $x$-axis in simulation. This alignment ensures that the robot's forward velocity command in simulation is geometrically consistent with the physical terrain layout.

\subsection{Rapid Test-Time Policy Training}

Here, test-time training means environment-specific fine-tuning in simulation before deployment. Starting from the pre-trained policy, we optimize on a reconstructed target terrain while preserving the same POMDP formulation (observation space, action space, termination criteria, and rewards). We investigate four fine-tuning strategies: \textbf{(1) Full Fine-tuning} updates all parameters end-to-end; \textbf{(2) Adapter Modules} insert lightweight zero-initialized adapters while freezing original weights; \textbf{(3) Residual Learning} adds a parallel zero-initialized network for additive action correction with a frozen base policy; \textbf{(4) Last Layer Fine-tuning} restricts updates to the final actor layer only. See \autoref{app:ttt_strategies} for details.

\section{Experiments}

In this section, we conduct a comprehensive evaluation of the \textit{TTT-Parkour} framework in both simulation and real-world scenarios. We deploy the policy to a suite of extremely challenging terrains designed to push the limits of humanoid locomotion, including wedges, stakes, boxes, trapezoids, and narrow beams. We aim to answer the following key questions:

\noindent \textbf{Q1. Adaptation \& Efficiency:} How much does terrain-specific fine-tuning improve our pre-trained policy, and how quickly does it adapt compared with training from scratch or joint multi-terrain fine-tuning?

\noindent \textbf{Q2. Reconstruction Quality \& Speed:} How do reconstruction fidelity and speed vary across different data sources, and is RGB-D necessary compared to RGB-only methods?

\noindent \textbf{Q3. Convergence Analysis:} What factors influence the sample efficiency and the required number of iterations for the test-time training process?

\begin{figure}[ht]
  \centering
  \includegraphics[width=0.9\linewidth]{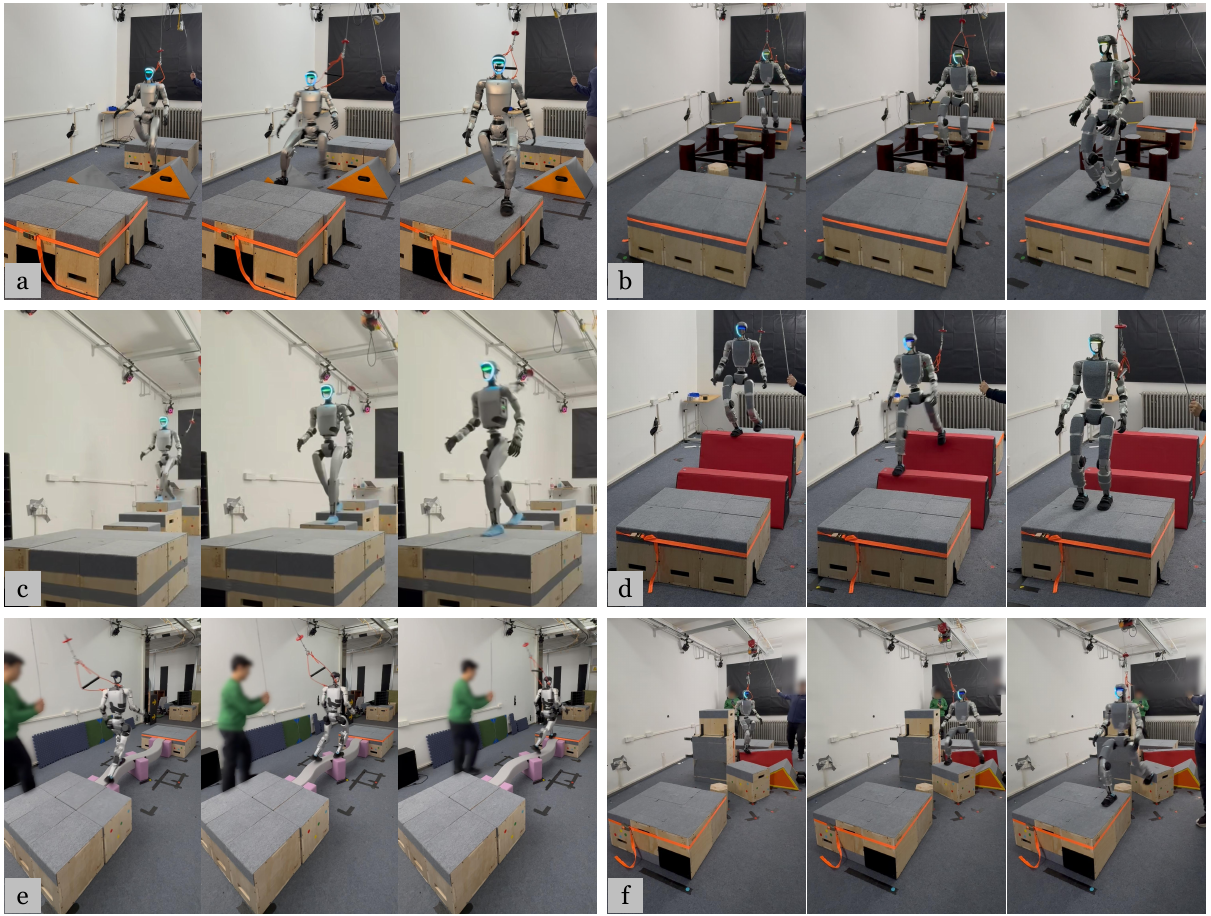}
  \caption{\textbf{Real-world experiments.} The robot successfully traverses extremely challenging terrains, including: (a) Wedges, (b) Stakes, (c) Boxes, (d) Trapezoids, (e) Narrow beams, and (f) Mixed terrain.}
  \label{fig:real-exp}
  \vspace{-6mm}
\end{figure}

\subsection{Experiment Configuration}

We use IsaacLab \cite{mittal2025isaac} with 4096 parallel Unitree G1 agents on an NVIDIA RTX 5090 (${<}4$s/iter). The deployed policy runs at 50 Hz on a Jetson Orin NX with a RealSense D435i. Pre-training requires approximately 5 days of wall-clock time on a single RTX 5090 GPU. We evaluate on 13 real-world terrains across 5 categories. Success rates use 5 real-world trials per terrain. \autoref{tab:sim_success} reports simulation mean$\pm$std over 3 evaluation seeds, with 1,000 trials per terrain per seed. See \autoref{app:exp_config} for full details.

\subsection{Traversability Analysis}

We compare TTT-1 with four baselines to evaluate adaptation performance and efficiency in simulation. The methods are defined as follows:

\noindent \textbf{(1) Pre-train}: The base policy is trained on large-scale procedurally generated terrains and deployed directly without adaptation.

\noindent \textbf{(2) Scratch-1}: A policy trained from random initialization directly on a single, specific reconstructed real-world terrain. 

\noindent \textbf{(3) Scratch-13}: A policy trained from scratch jointly on all 13 reconstructed test terrains for 30k iterations.

\noindent \textbf{(4) TTT-13}: A policy fine-tuned jointly on all 13 reconstructed test terrains until convergence at 1.5k iterations. 

\noindent \textbf{(5) TTT-1 (Ours)}: The proposed framework, which performs rapid test-time fine-tuning on a single, specific reconstructed mesh from a real-world environment.

\begin{figure}[t]
  \centering
  \begin{minipage}[t]{0.49\linewidth}
    \centering
    \includegraphics[width=\linewidth,trim=0 0 970.65bp 0,clip]{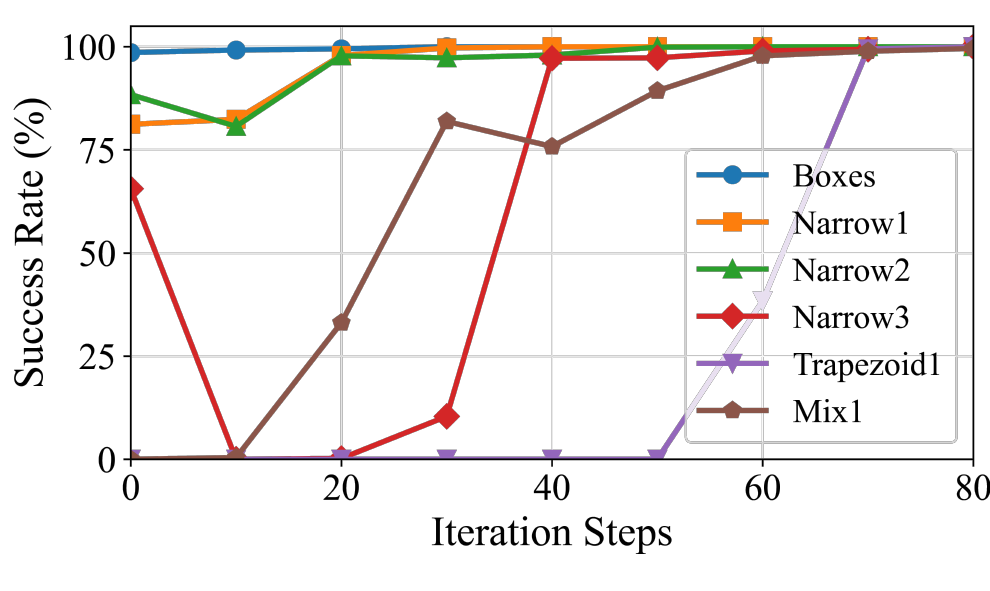}
    \par\small (a) Faster-converging terrains
  \end{minipage}\hfill
  \begin{minipage}[t]{0.49\linewidth}
    \centering
    \includegraphics[width=\linewidth,trim=481.88bp 0 481.88bp 0,clip]{figures/ttt-success.pdf}
    \par\small (b) Slower-converging terrains
  \end{minipage}
  \par\medskip
  \begin{minipage}[t]{0.49\linewidth}
    \centering
    \includegraphics[width=\linewidth,trim=963.77bp 0 0 0,clip]{figures/ttt-success.pdf}
    \par\small (c) Boston
  \end{minipage}\hfill
  \begin{minipage}[t]{0.49\linewidth}
    \centering
    \includegraphics[width=\linewidth]{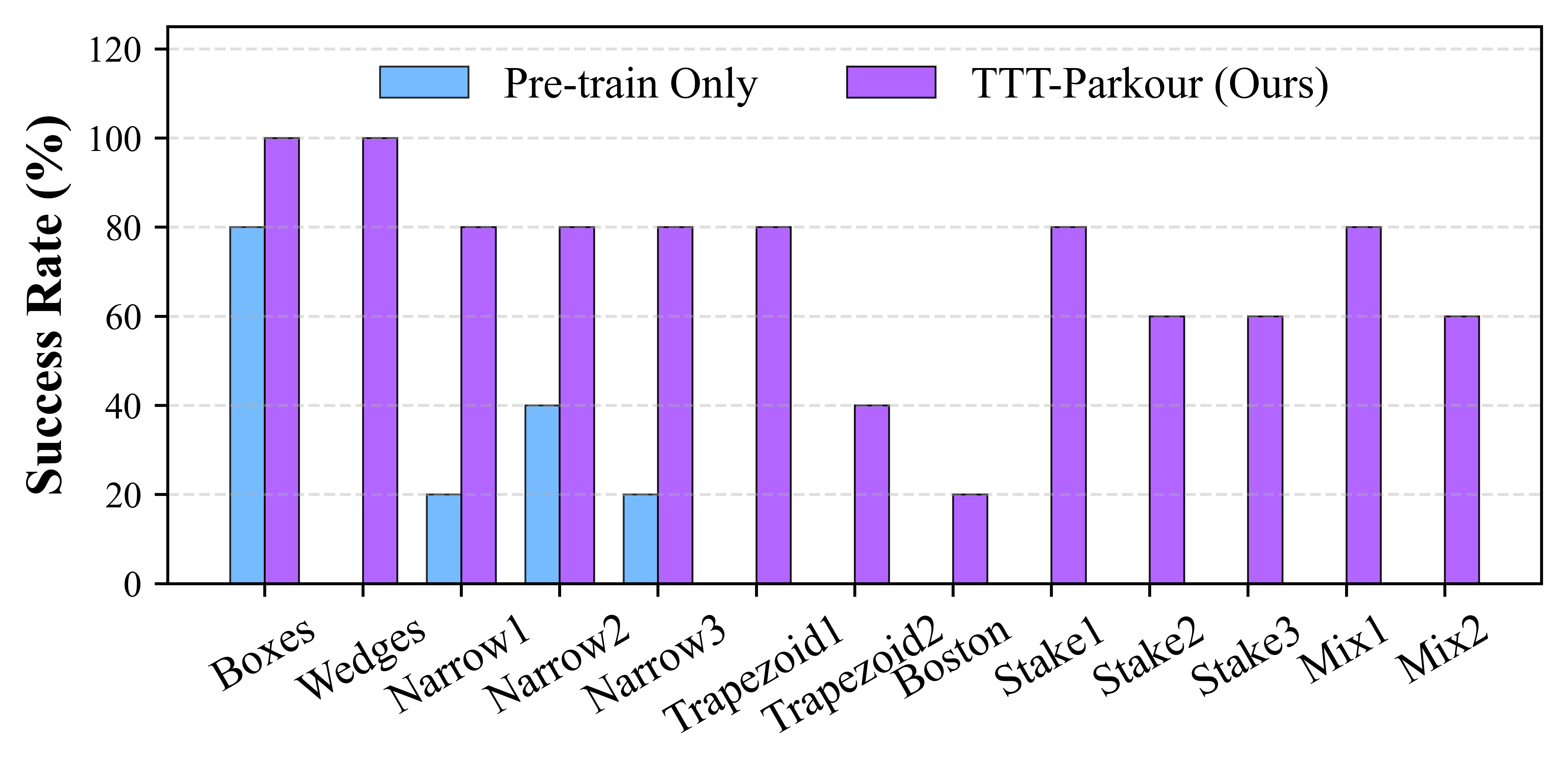}
    \par\small (d) Real-world evaluation
  \end{minipage}
  \caption{TTT-1 simulation success during fine-tuning on (a) Boxes, Narrow1--3, Trapezoid1, and Mix1; (b) Wedges, Stake1--3, Mix2, and Trapezoid2; and (c) Boston. (d) Real-world success of Pre-train and TTT-1 over 5 trials per terrain.}
  \label{fig:ttt-success}
\end{figure}

\autoref{fig:ttt-success}(a--c) shows the TTT-1 training progress. As reported in \autoref{tab:converge-steps}, 10 of the 13 terrains reach 97\% success within 120 iterations; Stake2, Trapezoid2, and Boston require 200, 240, and 560 iterations, respectively. Capture and reconstruction take 2 min 10 s (\autoref{tab:reconstruction_comparison}), and fine-tuning takes approximately 8 minutes for most tested terrains, giving a total of approximately 10 minutes. \autoref{tab:sim_success} compares the converged policies.

\textbf{\noindent(1)} TTT improves our pre-trained policy on the tested terrains. Most individual dimensions overlap with pre-training, while shapes, placements, and obstacle compositions differ (\autoref{tab:terrain_distribution}). These results establish gains over our base policy; stronger generalist policies may reduce the need for adaptation.

\textbf{\noindent(2)} Scratch-1 succeeds on two narrow-beam terrains after 25k iterations but fails on most others. Scratch-13 achieves 0\% on all terrains after 30k iterations. This supports pre-training as an effective initialization under our training setup.

\textbf{\noindent(3)} Converged TTT-13 and TTT-1 achieve comparable average success rates (99.56\% and 99.83\%), including Boston (98.93\% and 99.20\%). TTT-13 requires 1.5k iterations to converge, while TTT-1 reaches 97\% success on every terrain within 560 iterations (\autoref{tab:converge-steps}). The main benefit is faster terrain-specific specialization. Cross-terrain transfer depends on the fine-tuning terrain and is strongest among related configurations (\autoref{app:transfer}).

\begin{table}[t]
    \centering
    \caption{Simulation success rates (\%). Mean\textpm{}std over 3 evaluation seeds.}
    \label{tab:sim_success}
    \small
    \setlength{\tabcolsep}{5pt}
    \begin{tabular}{@{}lccccc@{}}
        \toprule
        Terrain & Pre-train & Scratch-1 & Scratch-13 & TTT-13 & TTT-1 \\
        & & (25k) & (30k) & (1.5k) & (Converged) \\
        \midrule
        Boxes & 98.27\textpm{}0.21\% & 0.00\textpm{}0.00\% & 0.00\textpm{}0.00\% & \underline{98.80\textpm{}0.46\%} & \textbf{100.00\textpm{}0.00\%} \\
        Wedges & 0.30\textpm{}0.30\% & 0.00\textpm{}0.00\% & 0.00\textpm{}0.00\% & \textbf{99.93\textpm{}0.12\%} & \underline{99.90\textpm{}0.00\%} \\
        Nar.1 & 82.70\textpm{}0.70\% & \textbf{100.00\textpm{}0.00\%} & 0.00\textpm{}0.00\% & 99.53\textpm{}0.21\% & \underline{99.97\textpm{}0.06\%} \\
        Nar.2 & 87.27\textpm{}1.24\% & \textbf{100.00\textpm{}0.00\%} & 0.00\textpm{}0.00\% & \textbf{100.00\textpm{}0.00\%} & \underline{99.97\textpm{}0.06\%} \\
        Nar.3 & 71.63\textpm{}4.08\% & 0.23\textpm{}0.12\% & 0.00\textpm{}0.00\% & \underline{99.20\textpm{}0.17\%} & \textbf{99.37\textpm{}0.15\%} \\
        Trap.1 & 0.00\textpm{}0.00\% & 0.00\textpm{}0.00\% & 0.00\textpm{}0.00\% & \textbf{100.00\textpm{}0.00\%} & \underline{99.97\textpm{}0.06\%} \\
        Trap.2 & 7.27\textpm{}1.21\% & 0.00\textpm{}0.00\% & 0.00\textpm{}0.00\% & \underline{99.57\textpm{}0.15\%} & \textbf{99.93\textpm{}0.06\%} \\
        Boston & 0.00\textpm{}0.00\% & 0.00\textpm{}0.00\% & 0.00\textpm{}0.00\% & \underline{98.93\textpm{}0.50\%} & \textbf{99.20\textpm{}0.00\%} \\
        Stake1 & 3.67\textpm{}0.40\% & 0.00\textpm{}0.00\% & 0.00\textpm{}0.00\% & \textbf{100.00\textpm{}0.00\%} & \underline{99.97\textpm{}0.06\%} \\
        Stake2 & 0.17\textpm{}0.06\% & 0.00\textpm{}0.00\% & 0.00\textpm{}0.00\% & \underline{99.00\textpm{}0.17\%} & \textbf{99.93\textpm{}0.06\%} \\
        Stake3 & 9.63\textpm{}1.32\% & 0.00\textpm{}0.00\% & 0.00\textpm{}0.00\% & \textbf{99.90\textpm{}0.17\%} & \underline{99.80\textpm{}0.17\%} \\
        Mix1 & 0.07\textpm{}0.06\% & 0.00\textpm{}0.00\% & 0.00\textpm{}0.00\% & \textbf{100.00\textpm{}0.00\%} & \underline{99.77\textpm{}0.23\%} \\
        Mix2 & 0.00\textpm{}0.00\% & 0.00\textpm{}0.00\% & 0.00\textpm{}0.00\% & \underline{99.43\textpm{}0.25\%} & \textbf{100.00\textpm{}0.00\%} \\
        \bottomrule
    \end{tabular}
\end{table}

\begin{figure}[b]
  \vspace{-4mm}
  \centering
  \includegraphics[width=\linewidth]{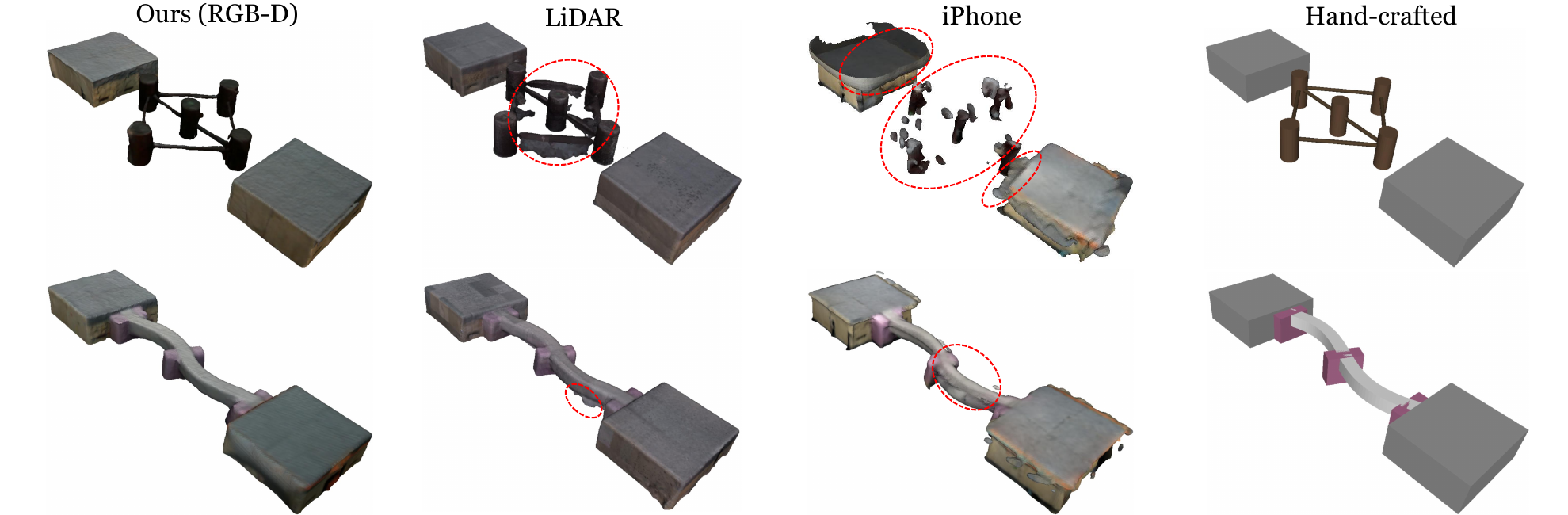}
  \caption{\textbf{Geometry comparison.} Our pipeline with RGB-D input balances reconstruction quality and efficiency when generating simulation-ready meshes. It maintains realistic geometric fidelity with significantly fewer artifacts compared to LiDAR and iPhone scans, where major artifacts are highlighted by red circles. Top row: Stake1, bottom row: Narrow1.}
  \vspace{-5mm}
  \label{fig:geo-exp}
\end{figure}

\begin{table}[t]
    \centering
    \caption{Comparison of different sources for reconstruction. Time indicates duration from data capture to final mesh.}
    \label{tab:reconstruction_comparison}

    {\small
    \begin{tabularx}{\textwidth}{l >{\centering\arraybackslash}X >{\centering\arraybackslash}X >{\centering\arraybackslash}X >{\centering\arraybackslash}X}
        \toprule
        \textbf{Methods} & \textbf{RGB-D} & \textbf{LiDAR Scanner} & \textbf{iPhone} & \textbf{Hand-crafted}\\
        \midrule
        \textbf{Pros}
        & Balanced quality and efficiency
        & Professional, best scale accuracy
        & Most accessible, fast acquisition
        & No artifacts \\
        \midrule
        \textbf{Cons}
        & Scale precision slightly lower than LiDAR
        & Terrain junction artifacts, tedious post-processing
        & Unstable reconstruction results, flying point artifacts
        & High manual effort, over-perfect geometry causes sim-to-real gap \\
        \midrule
        \textbf{Time} & \textbf{2 min 10 s} & 20 min & 4 min & 1 h \\
        \bottomrule
    \end{tabularx}
    }
    \vspace{-3mm}
\end{table}

\begin{table}[t]
  \centering\small
  \caption{Real-world success rates (\%) with different reconstruction sources.}
  \label{tab:recon-real-exp}
  \begin{tabular}{@{}lcccc@{}}
    \toprule
    Terrain & RGB-D & LiDAR & iPhone & Hand-crafted \\
    \midrule
    Stake1 & 80\% & 80\% & 0\% & 40\% \\
    Narrow1 & 80\% & 100\% & 80\% & 20\% \\
    Wedge & 100\% & 60\% & 20\% & 20\% \\
    Trapezoid1 & 80\% & 60\% & 20\% & 40\% \\
    \bottomrule
  \end{tabular}
\end{table}

\begin{table}[t]
  \centering\small
  \caption{Absolute relative scale errors (lower is better).}
  \label{tab:scale_errors}
  \begin{tabular}{@{}lccccc@{}}
    \toprule
    Terrain & RGB-D & LiDAR & iPhone & MapAnything & Pi3 \\
    \midrule
    Stake1 & \textbf{0.002} & \underline{0.016} & 0.074 & 0.383 & 0.863 \\
    Narrow1 & \underline{0.028} & \textbf{0.005} & 0.056 & 0.720 & 1.172 \\
    Wedge & \textbf{0.013} & \underline{0.015} & 0.027 & 0.944 & 0.423 \\
    Trapezoid1 & \textbf{0.018} & \underline{0.024} & 0.033 & 0.988 & 0.400 \\
    \bottomrule
  \end{tabular}
\end{table}

In the real-world experiments, we deploy the pretrained policy and terrain-specific policy obtained via test-time training \textbf{(TTT-1)} on each respective terrain. The success rates are shown in \autoref{fig:ttt-success}. As illustrated, the pre-trained general policy struggles significantly with geometries unseen during pre-training, failing completely (0\% success rate) on most of the challenging obstacles such as Wedges, Trapezoids, and Stakes. In contrast, \textit{TTT-Parkour} demonstrates robust adaptation capabilities, increasing success rates from near zero to more than 60\% on most complex terrains and achieving 100\% on Boxes and Wedges. This significant improvement validates that our test-time training pipeline effectively improves the robot's ability to adapt to diverse environments unseen during pre-training. Representative snapshots are shown in \autoref{fig:real-exp}. Real-world success rates are slightly lower than those in simulation. We attribute this sim-to-real gap to hardware instabilities (e.g., camera noise, actuator dynamics) and environmental mismatches. Unlike the static, rigid simulation, physical terrain elements often wobble or shift during robot interaction. These unmodeled dynamics can lead to failure in real-world tests. Additionally, discrepancies between the reconstructed geometry and the physical terrain still persist, further contributing to the performance gap.

\subsection{Mesh Reconstruction Analysis}

As shown in \autoref{fig:geo-exp}, \autoref{tab:reconstruction_comparison}, and \autoref{tab:scale_errors}, RGB-D achieves a favorable balance between reconstruction speed (2 min 10 s) and metric fidelity, achieving scale fidelity comparable to LiDAR while requiring no post-processing. Real-world deployment (\autoref{tab:recon-real-exp}) confirms this: RGB-D and LiDAR achieve comparable success rates, while iPhone scans and hand-crafted meshes underperform due to geometric artifacts and over-perfect geometry, respectively. Pi3 remains scale-ambiguous, while MapAnything can infer inaccurate metric scale on these terrains, validating the necessity of depth input for reliable sim-to-real transfer. Chamfer Distance and 95th-percentile Hausdorff Distance further quantify overall and local geometric errors (\autoref{tab:geometry_errors}, \autoref{app:mesh_recon}).

\subsection{Convergence Analysis}

As detailed in \autoref{tab:converge-steps} (Appendix), training from scratch fails to converge within 25k iterations on most terrains, with only the simpler narrow-beam terrains converging after 17k iterations, while TTT-1 reaches a 97\% success rate within 560 iterations across all 13 terrains. TTT-13 reaches comparable final performance after 1.5k iterations. Its slower adaptation may reflect reduced sampling per terrain; Trapezoid2 is an exception, potentially benefiting from transfer across related geometries.

\section{Conclusion and Limitations} 
\label{sec:conclusion}

We present \textit{TTT-Parkour}, a real-to-sim-to-real system combining efficient geometry reconstruction with environment-specific policy fine-tuning before deployment. It improves our pre-trained policy on challenging real terrains and provides faster specialization than joint multi-terrain fine-tuning.

\textbf{Limitations.} The current system requires manual handheld capture and offline fine-tuning on a workstation, taking approximately 10 minutes for most tested terrains. The fixed onboard camera cannot cover a full terrain scan; an actuated camera with autonomous scanning could reduce manual intervention. Adapted policies have limited transfer to unrelated terrains, and retaining the original policy does not establish preservation of its capabilities after fine-tuning. Benefits over stronger generalist policies remain untested. Real-world results use only 5 trials per terrain, and controlled geometry-perturbation tests and systematic evaluation on held-out terrain variations remain future work.

\clearpage
\acknowledgments{This project is supported by Tsinghua University Initiative Scientific Research Program.}

% No explicit bibliography style is required; corl_2026 loads corlabbrvnat.
\bibliography{references}

\clearpage
\appendix
\section*{Appendix}
\renewcommand{\thesection}{\Alph{section}}
\renewcommand{\thesubsection}{\Alph{section}.\arabic{subsection}}
\renewcommand{\thesubsubsection}{\Alph{section}.\arabic{subsection}.\arabic{subsubsection}}

\noindent\textbf{Table of Contents}
\vspace{2mm}

\noindent\begin{tabular}{@{}ll}
A & Policy Pre-training Details \dotfill \pageref{app:policy_details} \\
B & Mesh Reconstruction Analysis \dotfill \pageref{app:mesh_recon} \\
C & Convergence Analysis \dotfill \pageref{app:convergence} \\
D & Test-Time Training Strategies \dotfill \pageref{app:ttt_strategies} \\
E & Experiment Configuration \dotfill \pageref{app:exp_config} \\
F & Geometric Specifications of Real-World Terrains \dotfill \pageref{app:geo_spec} \\
G & Cross-Terrain Transfer \dotfill \pageref{app:transfer} \\
\end{tabular}
\vspace{4mm}

\section{Policy Pre-training Details}
\label{app:policy_details}

This section details the policy pre-training setup, including observation and action spaces, episode termination conditions, curriculum construction, AMP reference data, and reward formulation. The training grid consists of 20 columns and 10 rows, with 10 difficulty levels across the rows and 20 terrain columns where each terrain type occupies 4 columns.

\subsection{Observations, Actions, and Terminations}

The observation and action spaces are summarized in \autoref{tab:mdp_spec}. Stochastic noise is injected into all proprioceptive inputs during training to enhance sim-to-real robustness. Although the velocity-command interface includes the yaw command $\omega_z$, we set $\omega_z=0$ throughout the parkour task.

\begin{table}[H]
\centering
\small
\caption{Observation and action space specification.}
\label{tab:mdp_spec}
\renewcommand{\arraystretch}{1.2}
\setlength{\tabcolsep}{4pt}
\begin{tabular}{l l c l}
\toprule
\textbf{Space} & \textbf{Signal} & \textbf{Dim} & \textbf{Description} \\
\midrule
\multirow{7}{*}{\textbf{Actor Obs}}
  & Base Angular Velocity $\boldsymbol{\omega}_t$ & 3 & IMU gyroscope \\
  & Projected Gravity $\mathbf{g}_t$ & 3 & Gravity direction in base frame \\
  & Velocity Commands $\mathbf{c}_t$ & 3 & $(v_x, v_y, \omega_z)$ \\
  & Joint Positions $\mathbf{q}_t$ & 29 & 29-DoF joint angles \\
  & Joint Velocities $\dot{\mathbf{q}}_t$ & 29 & 29-DoF joint velocities \\
  & Last Action $\mathbf{a}_{t-1}$ & 29 & Previous target joint positions \\
  & Depth History $\mathcal{H}_t$ & $m{\times}H{\times}W$ & Strided depth frames (32\texttimes{}18 each) \\
\midrule
\textbf{Critic Obs} & + Base Linear Velocity $\mathbf{v}_t$ & 3 & Noise-free; critic only \\
\midrule
\textbf{Action} & Target Joint Positions $\mathbf{a}_t$ & 29 & Executed via PD control: $\boldsymbol{\tau}_t = k_p(\mathbf{a}_t - \mathbf{q}_t) - k_d\dot{\mathbf{q}}_t$ \\
\bottomrule
\end{tabular}
\end{table}

\textbf{Terminations:} Episodes terminate if any of the following occurs: (1) the episode times out; (2) the robot is stuck at the starting position for more than 4 seconds; (3) any body link contacts the ground; (4) the base orientation exceeds permissible thresholds.

\subsection{Policy Network}

The learning framework consists of three primary components: a CNN-based visual encoder, a Mixture-of-Experts (MoE) policy, and an Adversarial Motion Prior (AMP) discriminator. The visual encoder processes the strided depth history through a convolutional network, followed by a projection MLP to extract a visual embedding. This embedding is concatenated with the proprioceptive observations and fed into the MoE-based actor-critic network, where a gating mechanism dynamically weights the contributions of multiple expert networks to generate robust actions. Additionally, the AMP discriminator distinguishes between the actor's generated motions and the reference dataset. The detailed hyperparameters for these components are listed in \autoref{tab:net_arch_params}.

\begin{table}[H]
\centering
\small
\caption{Detailed hyperparameters for the network architecture.}
\label{tab:net_arch_params}
\renewcommand{\arraystretch}{1.2}
\setlength{\tabcolsep}{12pt}
\begin{tabular}{c|c}
\toprule
\textbf{Parameter} & \textbf{Value} \\
\midrule
CNN channels & [4] \\
CNN kernel sizes & [3] \\
CNN strides & [1] \\
CNN paddings & [1] \\
CNN pooling & MaxPool \\
Encoder projection units & [256, 256] \\
Encoder embedding dim & 128 \\
Encoder activation & ReLU \\
\midrule
Policy Type & MoE \\
Number of Experts & 4 \\
Gating hidden units & [256] \\
Expert hidden units & [256, 128, 64] \\
Policy activation & ELU \\
\midrule
Discriminator hidden units & [1024, 512] \\
Discriminator activation & ReLU \\
\bottomrule
\end{tabular}
\end{table}

\subsection{AMP and Rewards}

The AMP dataset's walking motions are generated by a Model Predictive Control (MPC) controller \cite{wholebodyhumanoidmpcweb} to ensure stable movement priors. During collection, we randomize the robot's command velocities every 5 seconds, sampling $v_x \in [0, 1.0]$ m/s, $v_y \in [-0.3, 0.3]$ m/s, and $\omega \in [-1.0, 1.0]$ rad/s. The complete dataset contains 57,979 frames recorded at 50 Hz, totaling 1,159.6 seconds.

We use the same reward formulation for pre-training and test-time training. The reward function comprises task $r_{\text{task}}$, regularization $r_{\text{reg}}$, safety $r_{\text{safe}}$, and AMP $r_{\text{AMP}}$ terms. Task reward uses dense velocity tracking derived from the goal vector, clipped to prevent reward hacking. Regularization penalizes terrain-edge footholds, energy consumption, and action rate. Safety terms enforce joint limits. AMP \cite{peng2021amp} encourages natural locomotion style using the MPC-generated reference dataset. Full reward specifications are provided in \autoref{tab:rewards_detail}.

The discriminator $D(\mathbf{S})$ receives short temporal sequences rather than single state pairs, enabling it to capture motion dynamics. It is trained with a least-squares (MSE) loss:
\begin{equation}
    L_D = \mathbb{E}_{\mathcal{M}}[(D(\mathbf{S}) - 1)^2] + \mathbb{E}_{\mathcal{P}}[(D(\mathbf{S}) + 1)^2],
\end{equation}
where $\mathcal{M}$ is the reference dataset and $\mathcal{P}$ is the policy rollout. Gradient penalty and weight decay are applied for training stability. The style reward fed back to the policy is:
\begin{equation}
    r_t = \max\left[ 0, 1 - 0.25(D(\mathbf{S}_t) - 1)^2 \right].
\end{equation}
Compared to binary cross-entropy and log-based rewards, this MSE loss and quadratic reward combination yields smoother, non-saturating gradients, improving convergence toward the reference motion manifold.

\begin{table}[H]
\centering
\caption{Detailed specification of reward terms, weights, and mathematical formulations.}
\label{tab:rewards_detail}
\footnotesize
\renewcommand{\arraystretch}{1.35}
\setlength{\tabcolsep}{6pt}
\resizebox{\linewidth}{!}{%
\begin{tabular}{l c l}
\toprule
\textbf{Reward Term} & \textbf{Weight} & \textbf{Mathematical Formulation} \\ 
\midrule
\multicolumn{3}{l}{\textit{\textbf{Task Reward}}} \\
Linear Vel Tracking ($v_{xy}$) & 2.0 & $\exp(-\| \mathbf{v}_{xy}^* - \mathbf{v}_{xy} \|^2 / 0.5^2)$ \\
Angular Vel Tracking ($\omega_z$) & 0.1 & $\exp(-(\omega_z^* - \omega_z)^2 / 0.5^2)$ \\
Heading Error & \textminus{}1.0 & $|\omega_z^*|$ \\
Don't Wait & \textminus{}0.5 & $\mathbb{I}(v_x^* > 0.3) \cdot (\mathbb{I}(v_x < 0.15) + \mathbb{I}(v_x < 0) + \mathbb{I}(v_x < -0.15))$ \\
Is Alive & 3.0 & $+1$ \\
Stand Still & \textminus{}0.3 & $(\|\mathbf{q} - \mathbf{q}_{default}\|_1 - 4.0) \cdot \mathbb{I}(\|\mathbf{v}^*\| < 0.15) \cdot \mathbb{I}(|\omega_z^*| < 0.15)$ \\
\midrule
\multicolumn{3}{l}{\textit{\textbf{Regularization Reward}}} \\
Edge Penetration & \textminus{}1.0 & $\sum_{i=1}^{|\mathcal{P}|} \|\mathbf{d}_i\| \cdot (\|\mathbf{v}_i\| + \epsilon)$ \\
Feet Air Time & 0.5 & $\min_{f} (t_{phase, f}) \cdot \mathbb{I}(\sum c_f = 1) \cdot \mathbb{I}(\|\mathbf{v}^*\| > 0.15)$ \\
Feet Slide & \textminus{}0.4 & $\sum_{f} \|\mathbf{v}_{xy, f}\| \cdot \mathbb{I}(c_f)$ \\
Joint Deviation (Hip) & \textminus{}0.5 & $\sum_{j \in \text{hips}} (q_j - q_{j, default})^2$ \\
Base Ang Vel (XY) ($L_2$) & \textminus{}0.05 & $\| \omega_{xy} \|^2$ \\
Joint Torques ($L_2$) & \textminus{}1.5e-7 & $\| \boldsymbol{\tau}_{legs} \|^2$ \\
Joint Acc ($L_2$) & \textminus{}1.25e-7 & $\| \ddot{\mathbf{q}} \|^2$ \\
Joint Vel ($L_2$) & \textminus{}1.0e-4 & $\| \dot{\mathbf{q}} \|^2$ \\
Action Rate ($L_2$) & \textminus{}0.005 & $\| \mathbf{a}_t - \mathbf{a}_{t-1} \|^2$ \\
Flat Orientation & \textminus{}3.0 & $\| \mathbf{g}_{xy}^{proj} \|^2$ \\
Pelvis Orientation & \textminus{}3.0 & $\| \mathbf{g}_{xy}^{proj, pelvis} \|^2$ \\
Feet Orientation & \textminus{}0.4 & $\sum_{f} \| \mathbf{g}_{xy, f}^{proj} \| \cdot \mathbb{I}(c_f)$ \\
Feet Height Error & \textminus{}0.1 & $\sum_{f} \sum_{p} \text{clip}(h_{f} - h_{terr, p} - 0.035, 0, 0.3) \cdot \mathbb{I}(c_f)$ \\
Feet Distance & 1.0 & $\exp(-\max(0, 0.12 - |y_{L}^R - y_{R}^R|) / 0.05) - 1$ \\
Energy Consumption & \textminus{}5.0e-5 & $\sum_{j} (\tau_j \dot{q}_j / k_j)^2$ \\
Freeze Upper Body & \textminus{}0.004 & $\|\mathbf{q}_{upper} - \mathbf{q}_{upper}^{default}\|_1$ \\
\midrule
\multicolumn{3}{l}{\textit{\textbf{Safety Reward}}} \\
Joint Pos Limits & \textminus{}1.0 & $\sum_{j} (\max(0, q_j - q_{j, max}) + \max(0, q_{j, min} - q_j))$ \\
Joint Vel Limits & \textminus{}1.0 & $\sum_{j} \max(0, |\dot{q}_j| - 0.9\dot{q}_{j, max})$ \\
Torque Limits & \textminus{}0.01 & $\sum_{j} \max(0, |\tau_j| - 0.8\tau_{j, max})^2$ \\
Undesired Contacts & \textminus{}1.0 & $\mathbb{I}(\text{count}(\text{collision}_{body \setminus feet}) > 0)$ \\
\midrule
\multicolumn{3}{l}{\textit{\textbf{AMP Reward}}} \\
AMP Style & 0.25 & $\max\left[ 0, 1 - 0.25(D(\mathbf{S}_t) - 1)^2 \right]$ \\
\bottomrule
\end{tabular}%
}
\end{table}
\clearpage

\section{Mesh Reconstruction Analysis}
\label{app:mesh_recon}

Our geometry reconstruction pipeline supports multiple input modalities, as illustrated in \autoref{fig:geo-exp} and \autoref{tab:reconstruction_comparison}. We use a RealSense D435i for RGB-D sensing, a Lixel K1 for LiDAR scanning, an iPhone 16 Pro with the \textit{3D Scanner App} for mobile scanning, and hand-crafted meshes generated with Python scripts using the \textit{Trimesh} library from manual measurements of the physical terrain.

Among these acquisition methods, RGB-D cameras use depth information to recover accurate physical scale and produce fewer artifacts than consumer-grade alternatives. This combination reduces the reconstruction-induced sim-to-real gap while retaining the reconstruction speed required for test-time parkour training.

LiDAR scanners provide professional-grade scale accuracy, but they are expensive and tend to introduce artifacts at terrain junctions and fine details. Their workflow is also labor-intensive: it requires multi-pass scanning, SLAM-based mapping, and substantial post-processing such as denoising and smoothing. Consumer devices such as the iPhone are accessible and fast, but their reconstructions can be inconsistent and noisy, including flying artifacts. Hand-crafted meshes are the least effective for transfer. Although they appear idealized to the human eye, they omit the surface irregularities and geometric noise present in the physical world, creating a severe domain mismatch.

To evaluate how reconstruction quality affects sim-to-real transfer, we deploy policies test-time trained on meshes from different sources in the real world, as reported in \autoref{tab:recon-real-exp}. While all policies converge to high success rates on their own terrain source in simulation, their real-world performance differs substantially. Across the four evaluated terrains, RGB-D achieves the strongest overall transfer, while LiDAR remains competitive and the iPhone and hand-crafted sources are less reliable. Hand-crafted terrains consistently underperform because of the domain mismatch above. For example, the iPhone-based reconstruction succeeds on the simpler \textit{Narrow1} terrain but fails on \textit{Stake1}, where artifacts are more severe in the complex multi-stake geometry. RGB-D and LiDAR obtain comparable real-world success rates. Although LiDAR slightly outperforms RGB-D on \textit{Narrow1} because of its scale accuracy, its hardware and processing costs make it less suitable for rapid test-time training.

We evaluate metric accuracy by registering each reconstructed point cloud to the hand-crafted reference with the measured physical scale and computing the absolute relative scale error. The results in \autoref{tab:scale_errors} show that RGB-D achieves scale fidelity comparable to industrial LiDAR while significantly outperforming RGB-only baselines. Pi3 remains scale-ambiguous, and MapAnything can infer inaccurate metric scale on these terrains, which supports using RGB-D input to reduce the sim-to-real gap.

\begin{table}[ht]
  \centering\small
  \caption{Reconstruction error: CD / HD95 (m; lower is better).}
  \label{tab:geometry_errors}
  \begin{tabular}{@{}lccccc@{}}
    \toprule
    Terrain & RGB-D & LiDAR & iPhone & MapAnything & Pi3 \\
    \midrule
    Stake1 & \textbf{0.050}/\textbf{0.078} & \underline{0.055}/\underline{0.149} & 0.115/0.194 & 0.115/0.227 & 0.195/0.343 \\
    Narrow1 & \underline{0.052}/\underline{0.076} & \textbf{0.047}/\textbf{0.071} & 0.135/0.232 & 0.115/0.187 & 0.211/0.379 \\
    Wedge & \textbf{0.029}/\textbf{0.036} & \underline{0.045}/\underline{0.057} & 0.048/0.067 & 0.198/0.344 & 0.236/0.396 \\
    Trapezoid1 & \textbf{0.029}/\textbf{0.036} & \underline{0.031}/\underline{0.080} & 0.151/0.247 & 0.210/0.353 & 0.209/0.356 \\
    \bottomrule
  \end{tabular}
\end{table}

\autoref{tab:geometry_errors} reports Chamfer Distance (CD) for overall geometric discrepancy and 95th-percentile Hausdorff Distance (HD95) for local errors. On Trapezoid1, RGB-D and LiDAR have similar CD (0.029 vs.\ 0.031 m), but different HD95 (0.036 vs.\ 0.080 m) and real-world success (80\% vs.\ 60\%). This suggests that local errors matter even when overall geometry is comparable.

\section{Convergence Analysis}
\label{app:convergence}

We evaluate the convergence efficiency of three training strategies: training from scratch, multi-terrain Test-Time Training (TTT-13), and terrain-specific TTT (TTT-1) across 13 terrains in simulation. Efficiency is quantified by the number of iterations required to reach a 97\% success rate. As detailed in \autoref{tab:converge-steps}, training from scratch requires substantially more iterations than TTT and fails to converge within 25k iterations on most terrains, with only the simpler narrow-beam terrains converging after 17k iterations.

\begin{table}[H]
  \centering\small
  \caption{Iterations to reach 97\% success. Boston's TTT-13 entry is a lower bound from the first 1k iterations.}
  \label{tab:converge-steps}
  \begin{tabular}{@{}lccc@{}}
    \toprule
    Terrain & Scratch-1 & TTT-13 & TTT-1 \\
    \midrule
    Boxes & \textgreater{}25k & 0 & \textbf{0} \\
    Wedges & \textgreater{}25k & 250 & \textbf{100} \\
    Nar.1 & 17k & 70 & \textbf{20} \\
    Nar.2 & 17k & 70 & \textbf{20} \\
    Nar.3 & \textgreater{}25k & 70 & \textbf{40} \\
    Trap.1 & \textgreater{}25k & 230 & \textbf{70} \\
    Trap.2 & \textgreater{}25k & \textbf{170} & 240 \\
    Boston & \textgreater{}25k & \textgreater{}1000 & \textbf{560} \\
    Stake1 & \textgreater{}25k & 350 & \textbf{100} \\
    Stake2 & \textgreater{}25k & 400 & \textbf{200} \\
    Stake3 & \textgreater{}25k & 260 & \textbf{100} \\
    Mix1 & \textgreater{}25k & 160 & \textbf{60} \\
    Mix2 & \textgreater{}25k & 270 & \textbf{120} \\
    \bottomrule
  \end{tabular}
\end{table}

Extending TTT-13 to 1.5k iterations yields 98.93\% mean success on Boston (\autoref{tab:sim_success}); \autoref{tab:converge-steps} retains the original lower bound for the first crossing of 97\%. Generally, TTT-13 exhibits slower adaptation compared to TTT-1. We attribute this to sample dilution: simultaneously optimizing for 13 terrains reduces the effective number of samples available for any specific terrain within a training batch, thereby slowing gradient updates for distinct geometries. However, \textit{Trapezoid2} presents a notable exception where TTT-13 converges faster than TTT-1 (170 vs. 240 iterations). We hypothesize that this terrain shares geometric similarities with the \textit{Narrow Beams}. Thus, multi-terrain TTT likely benefits from positive transfer, leveraging features learned from the \textit{Narrow Beams} to accelerate adaptation on \textit{Trapezoid2}.

\section{Test-Time Training Strategies}
\label{app:ttt_strategies}

\noindent \textbf{(1) Full Fine-tuning:} Updates all parameters of the policy network end-to-end, starting directly from the pre-trained checkpoint.

\noindent \textbf{(2) Adapter Modules:} Inserts lightweight adapters after each layer of the depth encoder and MLP. We freeze the original weights and optimize only the adapter modules. Adapter outputs are zero-initialized to preserve the original feature modulation at the start.

\noindent \textbf{(3) Residual Learning:} Adds a parallel network to learn an additive action correction ($\mathbf{a}_{total} = \mathbf{a}_{base} + \mathbf{a}_{res}$). The base policy is frozen, and the residual output layer is zero-initialized, effectively maintaining the original policy behavior initially.

\noindent \textbf{(4) Last Layer Fine-tuning:} Freezes the depth encoder and intermediate MLP layers, restricting updates exclusively to the final linear layer of the actor policy.

\begin{table}[H]
    \centering
    \caption{Comparison of different test-time training strategies: Iterations to reach a 97\% success rate.}
    \label{tab:ttt-strategies}
    \small
    \begin{tabular}{lcccc}
        \toprule
        \textbf{Terrains / Methods} & \textbf{Last Layer} & \textbf{Residual} & \textbf{Adapter} & \textbf{Full Fine-tuning (Ours)} \\
        \midrule
        Narrow1    & 120          & \underline{40} & \underline{40} & \textbf{20} \\
        Trapezoid1 & \textbf{60}  & \underline{70} & \textbf{60}   & \underline{70} \\
        Mix2       & \underline{200} & \textgreater{}1000     & 260           & \textbf{120} \\
        \bottomrule
    \end{tabular}
\end{table}

As shown in \autoref{tab:ttt-strategies}, \textbf{Full Fine-Tuning} achieves the most robust performance across the three tested terrains. Residual and Adapter methods require random initialization of new parameters, introducing sensitivity and extra convergence steps. Last Layer restricts the trainable parameter space, limiting adaptability. PEFT methods consistently underperform due to the significant domain shift demanding full parameter expressivity; since the primary bottleneck is physical simulation rather than gradient computation, PEFT provides negligible efficiency gains.

\section{Experiment Configuration}
\label{app:exp_config}

\subsection{Training Setup}

We utilize IsaacLab \cite{mittal2025isaac} powered by IsaacSim for high-fidelity physics simulation and policy training. All experiments are conducted on a workstation equipped with an NVIDIA RTX 5090 GPU, parallelized with 4096 humanoid robot agents, where each iteration takes less than 4 seconds. We leverage the NVIDIA Warp framework \cite{warp2022} to implement a GPU-accelerated ray-caster depth simulation. The robot employed is the 29-DoF Unitree G1. We pre-train the policy for 100,000 iterations, utilizing a history length of $h=8$ to capture temporal observations. Then we capture and reconstruct the scene for test-time training.

\subsection{Policy Deployment on Real Robot}

After environment-specific fine-tuning in simulation, we deploy the policy directly on the real robot without further policy updates. The policy is deployed on the robot's onboard NVIDIA Jetson Orin NX computer using ROS2, with the inference loop running at 50 Hz. For perception, we utilize the onboard Intel RealSense D435i camera operating at 60 Hz. The raw depth images are captured at a resolution of $480 \times 270$, downsampled to $64 \times 36$, and subsequently cropped to a $32 \times 18$ patch covering the center-bottom region to focus on the immediate terrain geometry.

\subsection{Tested Terrains}

Our experiment focuses on five terrain categories: wedges, stakes, boxes, trapezoids, and narrow beams. Across all experimental setups, the start and end platforms are constructed as large boxes measuring approximately $90 \text{ cm}$ in length, $80 \text{ cm}$ in width, and $35 \text{ cm}$ in height. All intermediate obstacles are arranged to require the robot to traverse them without touching the ground. In the \textbf{pre-training} stage, we employ procedural generation to create diverse variations of these five categories in simulation. To ensure robustness, we randomize the position, size, and shape of each obstacle. The simulation environment is organized as a $20 \times 10$ grid comprising $5$ distinct terrain categories, with each category occupying $4$ columns. Within each column, the $10$ rows follow a curriculum training strategy \cite{heess2017emergence}, where difficulty progressively increases from the first to the last row. Specifically, higher difficulty levels use larger gaps and smaller platform dimensions. For the \textbf{real-world} stage, we physically construct 13 distinct testing terrains spanning the aforementioned categories. Each terrain is characterized by randomized spatial arrangements and orientations to strictly challenge the robot's adaptability on unseen terrains.

\begin{table}[ht]
  \centering\small
  \caption{Pre-training and real-world terrain distributions. Dimensions are in meters.}
  \label{tab:terrain_distribution}
  \begin{tabularx}{\linewidth}{@{}lXX@{}}
    \toprule
    Terrain & Pre-training & Real-world test \\
    \midrule
    Boxes & Size 0.30--0.82; height 0.30--0.50; gap $\leq$0.35 & Height 0.40--0.50; gap 0.31--0.32 \\
    Wedges & Size 0.15--0.82; height 0.13--0.35; slope $\sim30^\circ$/60$^\circ$ & Height $\sim$0.31; slope 30$^\circ$ \\
    Narrow & Span 1.43--2.63; width 0.15--0.60; height 0.25--0.45; curved & Length $\leq$2.0; width 0.14--0.27; height 0.32--0.48; irregularly curved \\
    Stakes & Diameter 0.16--0.24; height 0.30--0.45; 5--7 piles & Diameter 0.18; height $\sim$0.40; irregular layouts \\
    Trapezoids & Top length 0.135--0.275; width 0.81--0.89; height 0.15--0.55 & Width up to 1.0; height 0.35--0.55 \\
    Mix1/2 & Single terrain family per training terrain & Cross-family obstacle compositions \\
    Boston & No direct counterpart & Irregular multi-level topology \\
    \bottomrule
  \end{tabularx}
\end{table}

\autoref{tab:terrain_distribution} compares the 200 pre-training terrains (40 configurations per group) with the test terrains. Most individual dimensions overlap, while irregular shapes, placements, and cross-family compositions introduce differences beyond scalar dimensions. This comparison describes coverage but does not quantify a statistical distance between the distributions.

\subsection{Evaluation Metrics}

We adopt the success rate as the primary metric. In each trial, the robot is initialized on the start platform and commanded with a constant forward linear velocity and zero angular velocity. A trial is recorded as a success if the robot successfully traverses the terrain and reaches the end platform without touching the ground or falling. We conduct 5 trials per real-world condition. For \autoref{tab:sim_success}, each simulation condition uses 1,000 trials per evaluation seed; mean$\pm$std summarizes 3 evaluation seeds of the trained policies. The transfer matrix uses the same trial count and reports means over 3 evaluation seeds. Evaluation varies initial positions and partially randomizes initial headings. This allows limited route adjustment but does not make the policies fully omnidirectional or establish generalization to perturbed terrain geometry.

\section{Geometric Specifications of Real-World Terrains}
\label{app:geo_spec}
\begin{figure}[ht]
  \centering
  \includegraphics[page=3, width=\textwidth]{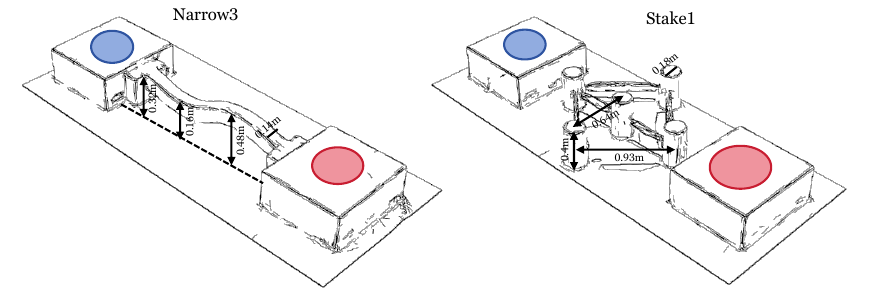}
    \caption{Detailed dimensions of the real-world terrains.}
    \label{fig:size1}
    
\end{figure}

\begin{figure}[ht]
  \centering
  \includegraphics[page=6, width=\textwidth]{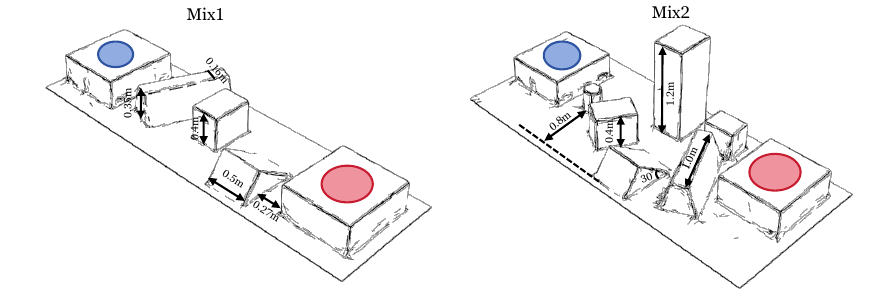}
  \caption{Detailed dimensions of the real-world terrains.}
      \label{fig:size2}
\end{figure}

\begin{figure}[t]
  \centering
  \includegraphics[page=7, width=\textwidth]{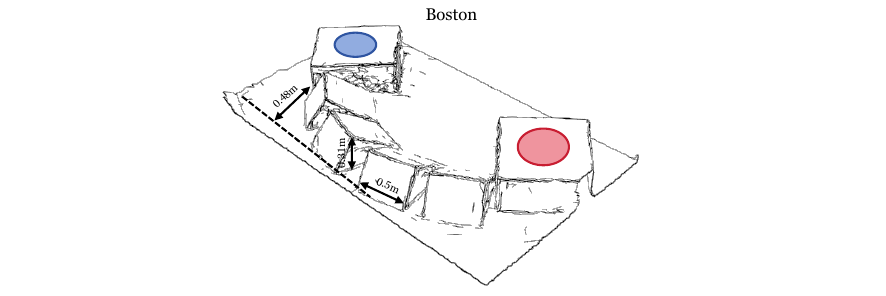}
  \caption{Detailed dimensions of the real-world terrains.}
      \label{fig:size3}
\end{figure}

We present the key dimensions of the terrains used in our real-world experiments, as shown in \autoref{fig:size1}, \autoref{fig:size2}, and \autoref{fig:size3}. In these figures, the red point denotes the starting platform $p_{\text{start}}$, and the blue point indicates the goal platform $p_{\text{end}}$. Several terrains feature extremely sparse or narrow geometries, posing significant challenges for the robot to secure stable footholds.

\begin{figure}[ht]
  \centering
  \includegraphics[page=1, width=\textwidth]{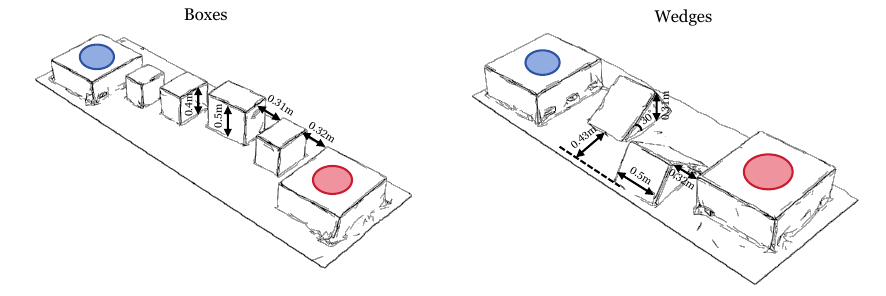}
\end{figure}

\begin{figure}[ht]
  \centering
  \includegraphics[page=2, width=\textwidth]{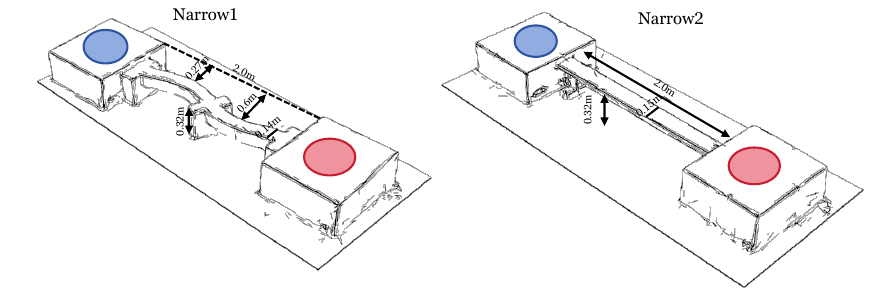}
\end{figure}

\begin{figure}[ht]
  \centering
  \includegraphics[page=4, width=\textwidth]{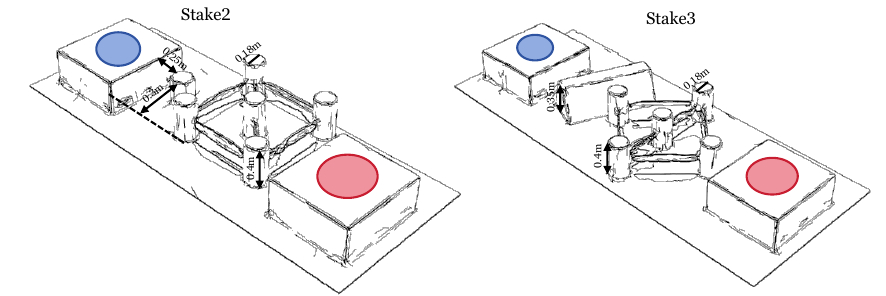}
\end{figure}

\begin{figure}[ht]
  \centering
  \includegraphics[page=5, width=\textwidth]{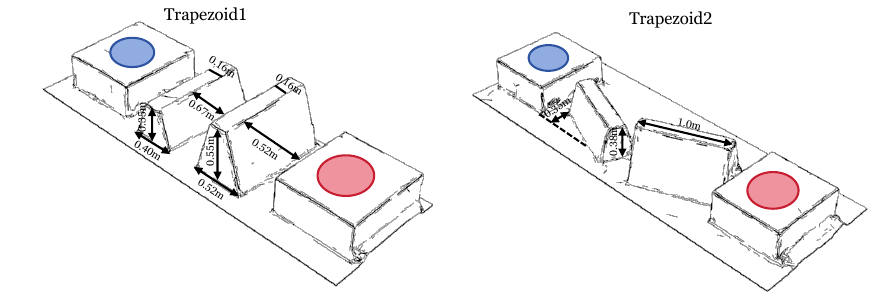}
\end{figure}

\clearpage
\section{Cross-Terrain Transfer}
\label{app:transfer}

\begin{figure}[ht]
  \centering
  \includegraphics[width=\linewidth]{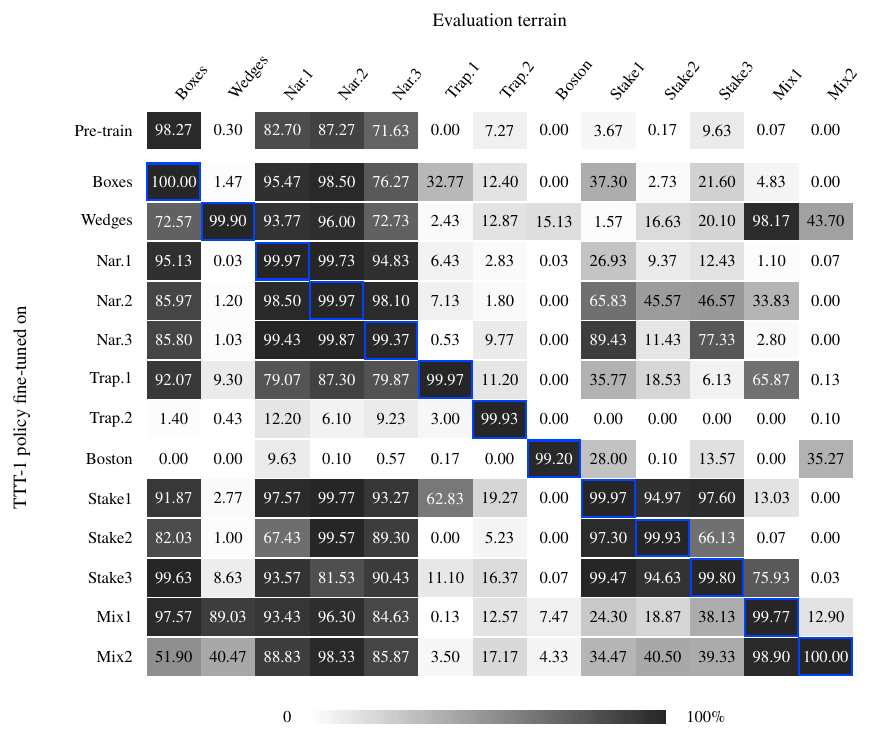}
  \caption{Cross-terrain simulation success (\%). Rows identify the fine-tuning terrain; columns identify the evaluation terrain. The top row is the pre-trained policy. Blue outlines mark evaluation on the fine-tuning terrain.}
  \label{fig:cross_terrain}
\end{figure}

We evaluate each terrain-specific TTT-1 policy on all 13 test terrains using 1,000 trials per cell per evaluation seed. \autoref{fig:cross_terrain} reports means over 3 evaluation seeds, with the pre-trained policy as a reference. Transfer is strongest among related configurations, especially narrow beams and stakes, and depends on the fine-tuning terrain. For example, the Stake1 policy achieves 94.97\% and 97.60\% on Stake2 and Stake3, whereas the Trapezoid2 policy performs poorly on most other terrains. We retain the original generalist policy for other environments and use adapted policies as on-demand specialists. These results do not establish retention across the full pre-training distribution.

\end{document}